\documentclass[11pt,a4paper]{article}
\usepackage[hyperref]{computel3}
\usepackage{times}
\usepackage{latexsym}

\usepackage{url}

\aclfinalcopy 

\usepackage{todonotes}

\usepackage{enumitem}
\usepackage{pgf-pie}
\usepackage{tabularx}
\usepackage{tikz}
\usepackage{tikz-qtree}
\usetikzlibrary{positioning,trees}

\newcounter{example}
\newenvironment{example}[1][]{\refstepcounter{example}}{}

\title{A Digital Corpus of St.~Lawrence Island Yupik}

\author{Lane Schwartz \\
  Department of Linguistics \\
  University of Illinois \\
  {\tt lanes@illinois.edu} \\\And
  Emily Chen \\
  Department of Linguistics \\
  University of Illinois \\
  {\tt echen41@illinois.edu} \\\AND
  Hyunji Hayley Park \\
  Department of Linguistics \\
  University of Illinois \\
  {\tt hpark129@illinois.edu} \\\And
  Edward Jahn \\
  {\tt ejahn3141@gmail.com} \\\And
  Sylvia L.R. Schreiner \\
  Linguistics Program \\
  Department of English \\
  George Mason University \\
  {\tt sschrei2@gmu.edu} \\}

\date{}

\begin{document}
\maketitle
\begin{abstract}
St.~Lawrence Island Yupik (ISO 639-3: \textit{ess}) is an endangered polysynthetic language in the Inuit-Yupik language family indigenous to Alaska and Chukotka.
This work presents a step-by-step pipeline for the digitization of written texts, and the first publicly available digital corpus for St.~Lawrence Island Yupik, created using that pipeline.
This corpus has great potential for future linguistic inquiry and research in NLP.
It was also developed for use in Yupik language education and revitalization, with a primary goal of enabling easy access to Yupik texts by educators and by members of the Yupik community.
A secondary goal is to support development of language technology such as spell-checkers, text-completion systems, interactive e-books, and language learning apps for use by the Yupik community.
\end{abstract}

\section{Introduction}

\begin{figure}[h]
\centering
\begin{tikzpicture}[point/.style={rectangle,draw,fill=black,inner sep=0pt,minimum size=0.001mm},]
%
\node[point] (IY1)  []  {};
\node[point] (IY2)  [right=5mm of IY1]  {};
\node[point] (I)    [above=20mm of IY2]  {};
\node[] (Inuit)  [right=27.1mm of I]  {Inuit languages};
\node[point] (S)    [below=10mm of I]    {};
\node[point] (Y)    [below=25mm of S]    {};
\node[point] (Y0)   [right=20mm of Y]    {};
\node[point] (Y1)   [above=15mm of Y0]    {};
\node[point] (Y2)   [below=10mm of Y1]    {};
\node[point] (Y3)   [below=10mm of Y2]    {};
\node[point] (Y4)   [below=10mm of Y3]    {};
\node[point] (Y5)   [above=9mm of Y1]    {};







\node[] (InuitBelow) [below left=0mm of Inuit] {};
\node[] (Greenland) [right=0mm of InuitBelow] {\textit{\footnotesize Greenland; Canada; Alaska}};

\node[] (Sirenik)  [right=27.1mm of S]  {Sirenik};
\node[] (SirenikBelow) [below left=0mm of Sirenik] {};
\node (SirenikLocation) [right=0mm of SirenikBelow] {\textit{\footnotesize Chukotka, Russia}};

\node[] (Siberian) [right=7mm of Y1] {\textbf{St.~Lawrence Island Yupik}};
\node[] (SiberianBelow) [below left=0mm of Siberian] {};
\node (SiberianLocation) [right=0mm of SiberianBelow] {\textit{\footnotesize Alaska; Chukotka, Russia}};

\node[] (Naukan) [right=7mm of Y2] {Naukan Yupik};
\node[] (NaukanBelow) [below left=0mm of Naukan] {};
\node (NaukanLocation) [right=0mm of NaukanBelow] {\textit{\footnotesize Chukotka, Russia}};

\node[] (Alaskan) [right=7mm of Y3] {Central Alaskan Yup'ik};
\node[] (AlaskanBelow) [below left=0mm of Alaskan] {};
\node (AlaskanLocation) [right=0mm of AlaskanBelow] {\textit{\footnotesize Western Alaska}};

\node[] (Alutiiq) [right=7mm of Y4] {Alutiiq Alaskan Yupik};
\node[] (AlutiiqBelow) [below left=0mm of Alutiiq] {};
\node (AlutiiqLocation) [right=0mm of AlutiiqBelow] {\textit{\footnotesize Southwest Alaska}};

\draw (IY1) -- (IY2);
\draw (I) -- (Y);
\draw (I) -- (Inuit);
\draw (S) -- (Sirenik);
\draw (Y) -- (Y0);
\draw (Y1) -- (Y4);
\draw (Y1) -- (Siberian);
\draw (Y2) -- (Naukan);
\draw (Y3) -- (Alaskan);
\draw (Y4) -- (Alutiiq);
\draw[dotted] (Y1) -- (Y5);

\end{tikzpicture}
\caption{\label{fig:tree}Inuit-Yupik language family \protect\cite{fortescue-etal-2010,Krauss:2011}}
\end{figure}
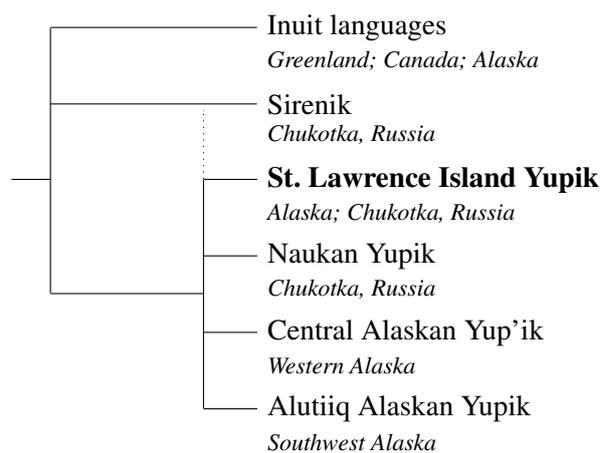

St.~Lawrence Island Yupik (ISO 639-3: \textit{ess}) is an endangered polysynthetic language in the Inuit-Yupik language family (see Figure~\ref{fig:tree}).
It is spoken on St.~Lawrence Island, Alaska and the Chukotka Peninsula of Russia.
This work presents the first publicly available digital corpus of written texts in St.~Lawrence Island Yupik, as well as the step-by-step process by which it was created.
We refer to this process as our \textit{digitization pipeline}, which can be readily adapted to any other language with any amount of written text.

The public release of the digital corpus has been coordinated with various stakeholders in the St.~Lawrence Island community, including the Native Village of Gambell, the Bering Strait School District, the Alaska Native Language Center at the University of Alaska Fairbanks, and Wycliffe Bible Translators.

The digital corpus is now available in plain-text format under the terms of the Creative Commons Attribution No-Commercial 4.0 International License at \url{https://github.com/SaintLawrenceIslandYupik/digital_corpus}.
Searchable PDF files are being archived at the Alaska Native Language Archive.
A mobile-friendly web-accessible version of the corpus will be subsequently developed to allow convenient on- or offline access to the corpus by members of the St.~Lawrence Island Yupik community.

\section{Goals for the Corpus\label{sec:goals}}

While the vast majority of St.~Lawrence Islanders born in or prior to 1980 are fluent L1 Yupik speakers \cite{Krauss:1980}, rapid language shift is underway among younger generations, especially in Russia where language shift is even further advanced \cite{Morgounova:2007}.
As a result, many members of the Yupik community have stated a desire for substantially strengthened Yupik instruction in the schools, ideally in the form of a Yupik language immersion program.
One obstacle to this, however, is that many Yupik-language texts as well as the pedagogical materials that were developed in the Soviet Union in the early 20th century \cite{Krauss:1971,Krupnik:2013} and in Alaska in the late 20th century \cite{Krauss:1971,Koonooka:2005} are not easily or broadly accessible.
Many materials are also archived at the Alaska Native Language Archive at the University of Alaska Fairbanks and at the Materials Development Center in the Gambell school.
Therefore, a primary goal for the development and release of this digital corpus is to strengthen opportunities for Yupik language revitalization and education by enabling easy access to existing Yupik-language texts by educators and by members of the Yupik community.
A related secondary goal is to support the development of language technologies such as spell-checkers, text-completion systems, interactive e-books, and language learning apps for use by the community.

\section{Digitization Pipeline \label{sec:digitization}}

We introduce in this section the digitization pipeline used in the creation of the digital corpus.
It consists of the following three steps, and can be easily replicated for other languages, since very few aspects of the pipeline were specially tailored to Yupik:
\begin{enumerate}[noitemsep]
    \item scanning
    \item image processing
    \item optical character recognition
\end{enumerate}
All of the texts that appear in the digital corpus are in UTF-8 plain-text format.

In the United States, Yupik is written using a Latin-derived orthography, while in Russia Yupik is written in a modified Cyrillic orthography.
The steps described in this section were applied to Yupik documents created in Alaska written in the Latin-derived Yupik alphabet.
A substantial amount of unscanned Cyrillic-orthography Yupik documents were gathered from Soviet libraries and archived at the Alaska Native Language Archive by \citet{Krauss:1971};
when the global COVID-19 pandemic situation once again allows for safe travel, we plan to scan and process these Cyrillic-orthography Yupik documents using essentially this same pipeline.

\subsection{Scanning\label{sec:scanning}}

During fieldwork visits to Gambell in 2017--2019, we identified and digitized a significant portion of the existing Yupik-language texts.
Priority was given to material most likely to be immediately useful in Yupik education efforts and in the development of Yupik language technologies, such as bilingual Yupik-English storybooks.

We gathered all Yupik language materials that could be found in the Gambell school library and Materials Development Center.
Most texts were scanned one page at a time using flatbed scanning equipment, while others were scanned using a sheet-fed scanner with an automatic page feeder feature in the Gambell school office.
There were a number of texts located at the Alaska Native Language Archive in Fairbanks that were not found in Gambell, and those were scanned on-site in early 2019.
Texts were scanned at a resolution of 600 DPI, and whenever possible saved in TIFF image format.

\subsection{Image Processing}

The raw TIFF image files were processed before optical character recognition was performed.
Any images that contained two physical pages were split into two separate files.
Next, images were deskewed, despeckled, and cropped.
In most cases, these steps were performed using Scan-Tailor,\footnote{\url{https://github.com/scantailor/scantailor}} an open source program designed for such image processing.
More recently, we have begun performing these image processing steps directly in ABBYY FineReader,\footnote{\url{https://pdf.abbyy.com}} a commercial application that we also use for performing optical character recognition.

\begin{figure}[!hb]
    \centering
    \includegraphics[scale=0.45]{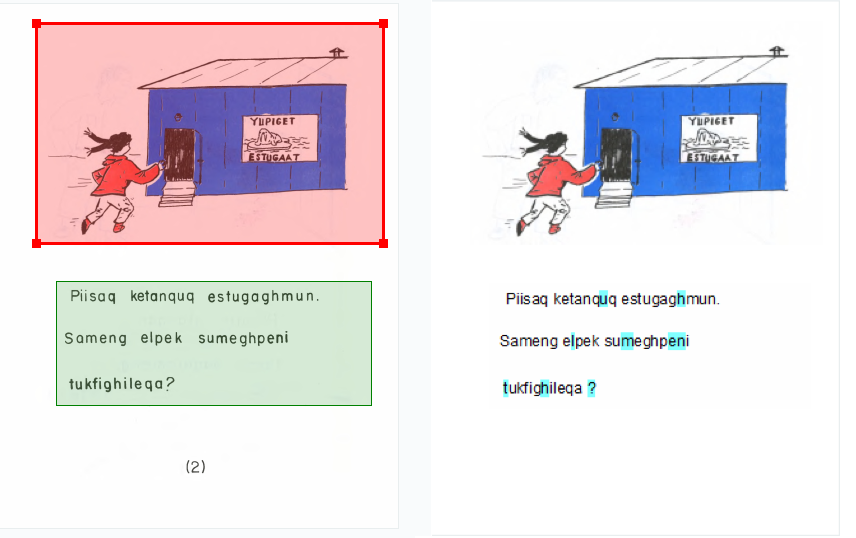}
    \caption{Regions of images and text are identified by ABBYY (left - red and green rectangles, respectively). Low-confidence characters are highlighted during OCR (right - cyan highlights).}
    \label{fig:abbyy}
\end{figure}

\begin{figure}[!h]
\centering
\fbox{\begin{minipage}{0.97\linewidth}
{\ttfamily\small
\fontfamily{cmtt}\selectfont
NAGATEK \\
\\
Ulimakat Nuum Agencym Mumiqhquqhyiqani , \\
Nuum Alaskami 99762 \\
Atughqaaluki Title VII-nem Maalghustun \\
akuzillghestun liinnaqfiganun Bureau of \\
Indian Affairenun .
}
\end{minipage}}
\caption{Sample plain text file of the Yupik front matter from the elementary reader \textbf{Nagatek} `\textit{Listen}'.}
\label{fig:yupik-frontmatter}
\end{figure}
\begin{figure}[!h]
\centering
\fbox{\begin{minipage}{0.97\linewidth}
{\ttfamily\small
\fontfamily{cmtt}\selectfont
Nagaten . \\
Nagaqughsigu-u ? \\
\\
Nakaa . \\
Sangaawa ? \\
\\
Esghaqaghhuqun tazigna . \\
\\
Maaten nagaqughaqa . \\
Enta aqfaatelta tazingavek . \\
\\
Hilikaptera .
}
\end{minipage}}
\caption{Sample plain text file of the Yupik content from the elementary reader \textbf{Nagatek} `\textit{Listen}'.}
\label{fig:yupik-content}
\end{figure}
\begin{figure}[!h]
\centering
\fbox{\begin{minipage}{0.97\linewidth}
{\ttfamily\small
\fontfamily{cmtt}\selectfont
LISTEN \\
\\
Written and Designed by Myra Poage \\
\\
Resource Staff/Translators \\
Raymond Oozevaseuk \\
Henry Silook \\
Linda S. Gologerqen \\
\\
Illustrated by Michael S. Apatiki \\
\\
A producation of the Nome Agency \\
Bilingual Education Resource Center, \\
P.O. Box 1108 \\
Nome, Alaska 99762 \\
for the Title VII Bilingual Education \\
Program of the Bureau of Indian Affairs \\
\\
Siberian Yupik \\
Printed at the GSA Printing Plant \\
P.O. Box 1612, Juneau, Alaska 99802 \\
\\
May 1975 \\
150 copies
}
\end{minipage}}
\caption{Sample plain text file of the English front matter from the elementary reader \textbf{Nagatek} `\textit{Listen}'.}
\label{fig:eng-frontmatter}
\end{figure}
\begin{figure}[!h]
\centering
\fbox{\begin{minipage}{0.97\linewidth}
{\ttfamily\small
\fontfamily{cmtt}\selectfont
1. Listen. \\
   Do you hear it? \\
2. No. \\
   What is it? \\
3. Look over there. \\
4. Now I hear it. \\
   Let’s run over there. \\
5. Helicopter.
}
\end{minipage}}
\caption{Sample plain text file of the English content from the elementary reader \textbf{Nagatek} `\textit{Listen}'.}
\label{fig:eng-content}
\end{figure}

\subsection{Optical Character Recognition}

%
\begin{table*}[!t]
    \centering
    \begin{tabular}{|r|r|r|r|r|r|r|}
     \cline{2-7}
       \multicolumn{1}{r|}{} &       \multicolumn{3}{c|}{\textbf{Yupik}}                      &             \multicolumn{3}{c|}{\textbf{English}}              \\ \hline
            \textbf{Corpus}  & \textbf{\# Sentences} & \textbf{\# Tokens} & \textbf{\# Types} & \textbf{\# Sentences} & \textbf{\# Tokens} & \textbf{\# Types} \\ \hline \hline
                  \textbf{Total} & 41,060 & 268,299 & 87,102 &        18,172 & 202,481 & 28,619      \\ \hline \hline
              Elementary Readers & 13,402 &  77,758 & 25,565 &         5,643 &  59,103 &  8,329      \\  
\textit{(front- \& back-matter)} &    818 &   2,364 &  1,053 &           962 &   7,903 &  2,429      \\ \hline
                 Oral Narratives &  9,818 &  64,696 & 24,883 &        10,374 & 120,194 & 12,516      \\ 
\textit{(front- \& back-matter)} &    275 &   1,149 &    760 &           886 &  12,909 &  4,581      \\ \hline
              Jacobson Exercises &    307 &     907 &    772 &           307 &   2,372 &    764      \\ \hline
                   New Testament & 16,440 & 121,425 & 34,069 &            \multicolumn{3}{c}{}       \\ \cline{1-4}
    \end{tabular}
    \caption{Counts of sentences, word tokens, and word types for texts included in the digital corpus. Some elementary readers and oral narratives include front-matter and/or back-matter. Note that the number of sentences (and tokens) in the Yupik and English corpora is not directly comparable - in the Yupik texts lines containing multiple sentences have been split apart and punctuation has been tokenized; in the English texts neither of these steps has yet been performed.}
    \label{tab:stats}
\end{table*}

Optical character recognition (OCR) is the process of converting an image into text.
As we began the process of scanning Yupik texts in 2016 and 2017, we first attempted to make use of the open source Tesseract\footnote{\url{https://github.com/tesseract-ocr/tesseract}} OCR software to convert the scanned images into text.
While Tesseract models can be trained for new languages, such training requires existing digitized texts.
This resulted in a bootstrapping dilemma; without existing Yupik digital texts, we could not train Tesseract models for Yupik.

After poor initial results with Tesseract, we made the decision to switch to ABBYY FineReader (hereafter ABBYY), a state-of-the-art commercial OCR application, for converting our processed image files to plain text.
This software was available to us through our respective libraries at the University of Illinois and George Mason University.
ABBYY FineReader includes pre-trained OCR models for the broader Inuit-Yupik language family in both Latin and Cyrillic orthographies.
Initial work was performed using ABBYY version 12, with later work performed using ABBYY version 14.
Unlike our early attempts with Tesseract, OCR quality in both versions of ABBYY FineReader was acceptable.

To begin an ABBYY OCR project, all of the TIFF images associated with a document are imported, analyzed, and partitioned into regions that contain text and regions that contain images or illustrations.
These regions must be verified, and can be corrected where necessary.

For each text region, we use ABBYY's built-in support for texts written in ``Eskimo Latin'' to perform OCR, and rely on this language setting for all Latin-orthography Yupik documents.
%
%
We have observed good OCR results even for documents that have deteriorated somewhat over time.
ABBYY will nevertheless identify low-confidence characters in the recognized text, and present them to the user for validation.
For each section of text that includes one or more low-confidence characters, the TIFF image associated with that section is presented to the user beside a pre-populated text box, in which low-confidence characters are highlighted.
The user then confirms or corrects each of these characters.
These aspects of image analysis and OCR are shown in Figure~\ref{fig:abbyy}.

Each fully OCR'd document is then saved in three file formats:
\begin{itemize}
    \item Microsoft Word DOCX
    \item searchable PDF/A
    \item UTF-8 plain text
\end{itemize}
The Microsoft Word documents will be shared with instructional staff at the St.~Lawrence Island schools.
Searchable PDF/A files will be archived at the Alaska Native Language Archive, and plain text files are included in our digital corpus.

Lastly, the plain text files are subsequently separated and saved as four individual files.
The first file contains any Yupik-language front-matter and back-matter, including title page, table of contents, and appendices (Figure~\ref{fig:yupik-frontmatter}).
The second file contains the main body of the Yupik text, excluding any front-matter and back-matter (Figure~\ref{fig:yupik-content}).
The third file contains the English front-matter and back-matter, if any (Figure~\ref{fig:eng-frontmatter}), and the fourth file contains the English translation of the main body of the text, if any (Figure~\ref{fig:eng-content}).
Furthermore, each sentence of a file appears on its own line, a blank line is used to delimit paragraphs, and punctuation marks are separated out from each line of text.
We formatted the files of the digital corpus in this way to ensure that there not only exists a record of the full text, but to also facilitate any NLP work that uses this corpus as a source of data.
Separating each text into four individual files enables researchers to easily access the desired data which typically does not include front and back matter.
The formatting of each individual file is likewise intended to facilitate text processing.

\nocite{YupikBible}

\section{The Digital Corpus of Yupik}

To date, we have digitized most of the existing Yupik-language texts using the digitization pipeline introduced herein.
%
We have scanned 90 mostly comb-bound Yupik elementary readers, 7 collections of Yupik oral narratives, the end-of-chapter exercises from the \newcite{Jacobson:2001} grammar, and 14 collections of Yupik-language hymns and other religious texts.
Table \ref{tab:stats} summarizes the distribution of sentences, word types, and word tokens across the current digital corpus.
We describe each type of text in the following sections.

\subsection{Elementary-Level Readers\label{sec:readers}}

In the 1970s, a set of elementary-level readers were developed by the Nome Agency Bilingual Education Resource Center at the Bureau of Indian Affairs and by the Alaska Native Language Center at the University of Alaska Fairbanks.
In the 1980s, additional readers were developed by the Bering Strait School District's Bilingual Materials Development Center (MDC) at the Gambell School on St.~Lawrence Island.
In the early 1990s, a series of five bilingual Yupik-English readers was planned for use by the St.~Lawrence Island Schools in grades 4-8 \cite{Apassingok:1993}.
Only the first three books in the series \cite{Apassingok:1993,Apassingok:1994,Apassingok:1995} were actually produced.

To date, 90 of these elementary-level readers have been scanned.
Of those, 68 have been fully digitized and are included in the digital corpus, including the bilingual grade 4-6 readers.
Processing of the remaining 22 elementary-level readers is ongoing.
As seen in Figures~\ref{fig:sentence-distributions} and \ref{fig:type-distributions}, while the elementary-level readers comprise nearly half of the sentences in the digital corpus, they contribute far fewer word types.
This is to be expected, given the nature of these texts;
since they were originally intended for language learning, one would expect them to frequently repeat words.

\subsection{Oral Narratives\label{sec:narratives}}

In the late 1970s, two books collecting St.~Lawrence Island legends were produced by the National Bilingual Materials Development Center at the University of Alaska Anchorage \cite{Slwooko:1977,Slwooko:1979}.
In the 1980s, a set of Yupik oral narratives were recorded on cassette tape, a subset of which were transcribed, translated, and collected into a series of three books constituting the Lore of St.~Lawrence Island collection \cite{Apassingok:1985:Vol1,Apassingok:1987:Vol2,Apassingok:1989:Vol3}. 
%
%
In the late 1990s, a collection of oral narratives were recorded in Savoonga, Alaska as part of fieldwork conducted by Japanese linguist Kayo Nagai. 
Transcriptions of these narratives, along with interlinear glosses and free translations, were later published in book form \cite{Nagai:2001}.
In the early 2000s, a 20th-century collection of Yupik oral narratives from Chukotka was transliterated from Cyrillic into the Latin Yupik orthography and published with English translations \cite{Koonooka:2003}.
The oral narratives in these collections include short stories, legends and folktales orally narrated by Yupik elders.

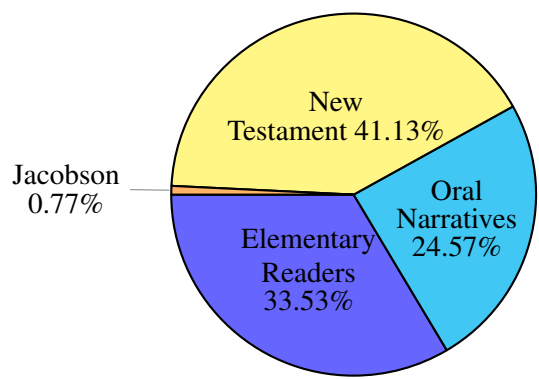
\begin{figure}[!h]
\vspace*{-3mm}
\centering
    \begin{tikzpicture}[scale=0.8]
     \pie [rotate=180,text=inside,outside under=10,no number] 
     {33.53/{Elementary \\ Readers \\ 33.53\%}, 24.57/{Oral \\ Narratives \\ 24.57\%}, 41.13/{New \\ Testament 41.13\%}, 0.77/{Jacobson \\ 0.77\%}}
    \end{tikzpicture}
\caption{Distribution of total Yupik sentences per collection, excluding front- \& back-matter and English content.}
\label{fig:sentence-distributions}
\end{figure}

\begin{figure}[!h]
\vspace*{-5mm}
\centering
    \begin{tikzpicture}[scale=0.8]
     \pie [rotate=180,text=inside,outside under=10,no number]
     {29.97/{Elementary \\ Readers \\ 29.97\%}, 29.17/{Oral \\ Narratives \\ 29.17\%}, 39.95/{New \\ Testament \\ 39.95\%}, 0.91/{Jacobson \\ 0.91\%}}
    \end{tikzpicture}
\caption{Distribution of total Yupik word types per collection, excluding front- \& back-matter and English content.}
\label{fig:type-distributions}
\end{figure}
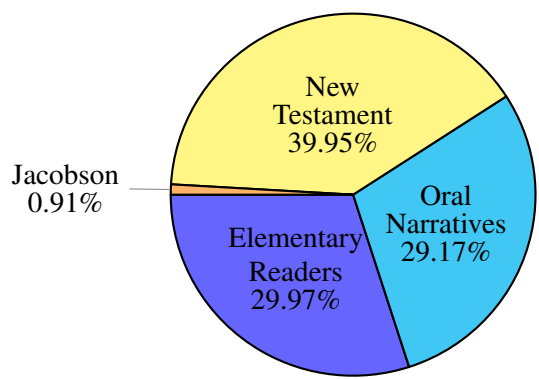

To date, five of these collections of Yupik oral narratives have been fully digitized and are included in the digital corpus, while processing of the remaining two \cite{Slwooko:1977,Slwooko:1979} is ongoing.
As seen in Figures~\ref{fig:sentence-distributions} and \ref{fig:type-distributions}, the oral narratives contribute approximately one third of the sentences and word types in the digital corpus.


\subsection{\protect\newcite{Jacobson:2001} End-of-Chapter Exercises\label{sec:eoc}}

The most thorough source of language documentation for Yupik is the grammar of \newcite{Jacobson:2001}.
The grammar is written at the level of an undergraduate college text, and appears to be designed for an audience of L1 Yupik speakers studying their own language at the college level.
Chapters 3--17 of the grammar each include end-of-chapter sample Yupik sentences.
These sentences are designed for the reader to practice the aspects of Yupik grammar presented in each respective chapter by translating the sentences into English.
These end-of-chapter sample sentences have been fully digitized and are included in the digital corpus, though they comprise only a small portion of the corpus as seen in Figures~\ref{fig:sentence-distributions} and \ref{fig:type-distributions}.
As part of our Yupik research, we elicited English reference translations of the Jacobson sample sentences which we also include in the digital corpus.

\subsection{Religious Texts\label{sec:religious}}

A Yupik translation of the New Testament was published in 2018, completing a nearly 60-year collaborative translation project by Wycliffe Bible Translators and Yupik translators on St.~Lawrence Island.
%
%
%
%
There are 14 additional Yupik religious texts (including a collection of hymns) from the Alaska Native Language Archive that we have scanned but not yet fully processed.
%

\section{Corpus's Potential for Linguistic Study}

Given Yupik's status as an understudied language, there is no doubt much to be learned linguistically from analyzing this digital corpus.
While the corpus has yet to be annotated, preliminary work has yielded several remarkable facts of the language that were not known to us previously, particularly with respect to its morphology.

The morphology of Yupik is perhaps one of the more well-documented aspects of the language.
The Yupik lexicon is broadly composed of three types of morphemes: roots, derivational morphemes, and inflectional morphemes.
Since Yupik is strictly suffixing with the exception of one prefix, words typically have the following form:
\begin{center}
\textit{root}-\textit{derivational morpheme(s)}-\textit{inflectional morpheme}
\end{center}
Most roots are nominal or verbal, such as \textit{alquutagh-} `spoon' and \textit{qepghagh-} `to work' respectively.
This results in four types of derivational morphemes:

\begin{itemize}
    \item N$\rightarrow$N which suffix to nominal roots and yield nominal stems
    \item N$\rightarrow$V which suffix to nominal roots and yield verbal stems
    \item V$\rightarrow$V which suffix to verbal roots and yield verbal stems
    \item V$\rightarrow$N which suffix to verbal roots and yield nominal stems
\end{itemize}

Upon suffixation, there are several morphophonological processes that can occur depending on the phonological shape of the root and the morpheme being suffixed.
For instance, in (\ref{ex:atkuperaq}), suffixing the derivational morpheme \textit{-peragh-} results in the deletion of root-final consonant \textit{-g-} in the root \textit{atkug-}.

\vspace{6pt}
\noindent \begin{example}
\begin{tabularx}{\columnwidth}[!h]{@{}*5l@{}}
(\theexample) & \textbf{atkuperaq} \\
              & atkug-peragh-\O \\
              & parka-makeshift-\textsc{abs}.sg \\
              & `\textit{makeshift parka}' \citep[p.661]{Badten:2008} \\
\label{ex:atkuperaq}
\end{tabularx}
\end{example}

The \citet{Badten:2008} Yupik-English dictionary comprehensively documents all of the morphemes that have been identified to date, while the \citet{Jacobson:2001} reference grammar and \citet{deReuse:1994} overview the known ordering constraints (e.g. V$\rightarrow$V derivational morphemes may only suffix to verbal roots and stems) and all of the morphophonological processes that can occur upon suffixation.
The corpus, however, has demonstrated several exceptions to this documentation.

For example, the derivational morpheme \textit{-pag-} is an augmentive V$\rightarrow$V suffix meaning `to V intensively, excessively'.
As such, it is attested to suffix to verbal stems only.
One would thus expect it to yield the word seen in (\ref{ex:qepghaghpagtuq}) but not in (\ref{ex:alquutaghpaget}), where the stem \textit{alquutagh-} `spoon' is a nominal stem.

\vspace{6pt}
\noindent \begin{example}
\begin{tabularx}{\columnwidth}[!h]{@{}*5l@{}}
(\theexample) & \textbf{qepghaghpagtuq} \\
              & qepghagh-pag-tu-q \\
              & work-\textsc{aug-ind.intr}-3sg \\
              & `\textit{he worked hard}' \citep[p.659]{Badten:2008} \\
\label{ex:qepghaghpagtuq}
\end{tabularx}
\end{example}

\noindent \begin{example}
\begin{tabularx}{\columnwidth}[!h]{@{}*2l@{}}
(\theexample) & \textbf{alquutaghpaget} \\
              & alquutagh-pag-et \\
              & spoon-\textsc{aug-abs}.pl \\
              & `\textit{heaping tablespoons}' \citep[p.103]{Nagai:2001} \\
\label{ex:alquutaghpaget}
\end{tabularx}
\end{example}
\vspace{-6pt}

Nevertheless, (\ref{ex:alquutaghpaget}) is a valid, attested form in our digital corpus, and \textit{-pag-} frequently appears suffixed to other nominal stems as well (\textit{sagneghpaget
} `large bowls', \textit{neqekrangllaghpagni} `great smell of bread', \textit{suupeliighpagni} `great smell of stew' \citep{Apassingok:1993}).
This suggests that perhaps \textit{-pag-} is not only a V$\rightarrow$V suffix, but also an N$\rightarrow$N suffix, which is not attested in the existing documentation.

A second interpretation is that there exist fewer constraints on morpheme ordering than previously believed, which would permit V$\rightarrow$V suffixes to affix to nominal roots.
This is also supported by substantial evidence of verbal roots being inflected for nominal inflectional morphemes in our corpus, as seen in (\ref{ex:yuvghiiq}).

\defcitealias{Aghnaaneq_Neghsaq_Teghigniqelghii}{NABERC, 1975}
\vspace{6pt}
\noindent \begin{example}
\begin{tabularx}{\columnwidth}[!h]{@{}*2l@{}}
(\theexample) & \textbf{yuvghiiq} \\
              & yuvghiiq-\O \\
              & examine-\textsc{abs}.sg \\
              & `\textit{look!}' \citepalias{Aghnaaneq_Neghsaq_Teghigniqelghii} \\
\label{ex:yuvghiiq}
\end{tabularx}
\end{example}
\vspace{-6pt}

In the same way one would not expect the V$\rightarrow$V suffix \textit{-pag-} to suffix to a nominal root, one would not expect a verbal root to inflect for nominal inflectional endings.
In this way, the digital corpus has opened up a rich area of inquiry.

The digital corpus also speaks to the influence of Yupik's oral tradition.
Since many of the texts in our corpus were originally oral narrations, there is considerable speaker variation, which has resulted in transcriptions of morphemes that differ greatly from their attested forms in the \citet{Badten:2008} dictionary.
These variations are apparent in the morphophonology as well, particularly in regards to allomorphy, as seen in (\ref{ex:maklaguugut}).

\vspace{6pt}
\noindent \begin{example}
\begin{tabularx}{\columnwidth}[!h]{@{}*2l@{}}
(\theexample) & \textbf{maklaguugut} \\
              & maklagu-u-gu-t \\
              & bearded.seal.intestines-be-\textsc{ind.intr}-3pl \\
              & literal trans. \textit{they are bearded seal's intestines} \\
              & \citep[p.191]{Nagai:2001} \\
\label{ex:maklaguugut}
\end{tabularx}
\end{example}

\noindent \begin{example}
\begin{tabularx}{\columnwidth}[!h]{@{}*2l@{}}
(\theexample) & \textbf{maklagunguut} \\
              & maklagu-ngu-u-t \\
              & bearded.seal.intestines-be-\textsc{ind.intr}-3pl \\
              & literal trans. \textit{they are bearded seal's intestines} \\
\label{ex:maklagunguut}
\end{tabularx}
\end{example}
\vspace{-6pt}

Whereas the word form in the digital corpus is \textit{maklaguugut}, the word form predicted by the attested morphophonological processes of Yupik is \textit{maklagunguut}, seen in (\ref{ex:maklagunguut}).
Detailed analysis of word forms such as these would contribute to an increased understanding of Yupik morphophonology, and remedy gaps in the existing documentation.

Our digital corpus thus offers many possibilities for future research in and documentation of Yupik morphology.
It would not only facilitate studies on morpheme ordering constraints and morphophonological variation, but would also allow for the potential discovery of novel, previously unattested morphemes.
Beyond the level of the word, the corpus will be of use for the study both of morphemes in context and of phenomena at the level of the sentence or the discourse. For example, Yupik boasts a large number of morphemes with meanings related to tense, aspect, mood, and modality. The meanings and uses of these morphemes are not well documented to begin with; in addition, their use often depends on factors outside of the word or sentence they are in. Sentential and discourse-level context is essential for understanding and analyzing many other syntactic and semantic phenomena, as well.

\section{Corpus's Potential for NLP Research\label{sec:nlp-potential}}

Many polysynthetic languages, such as Yupik, are low-resource and under-researched within the field of NLP.
The availability of the digital corpus for Yupik now enables researchers to utilize a written dataset that was otherwise inaccessible.
For our own purposes the digital corpus has had an immediate impact on two of our projects related to NLP, that is, our ongoing development of a morphological analyzer and a dependency treebank for Yupik.

Given Yupik's rich morphology, the implementation of a morphological analyzer is an essential step in the development of more complex language technologies.
While two iterations of a rule-based morphological analyzer have already been implemented \citep{Chen:2018,Chen:2020}, neither achieve full coverage and provide an analysis for all input items.
The digital corpus, however, offers a means of understanding the shortcomings of our existing analyzers.

We have already begun a detailed error analysis of the words in the corpus that cannot be analyzed, and are working on identifying the prominent patterns in these errors.
For instance, as described in the previous section, our corpus has demonstrated that constraints on morpheme ordering are perhaps more lax than has been initially documented.
Knowing this allows us to appropriately modify the analyzer to take this phenomena into account.
All of our findings from studying the digital corpus will subsequently be used to improve the existing analyzers.

The digital corpus is also currently being used to create the first Universal Dependencies (UD) \cite{nivre-etal-2016-universal}  treebank for Yupik.
UD provides a crosslingual framework for consistent annotation of dependency grammars across different natural languages.
However, the framework has not often been utilized for annotating polysynthetic languages like Yupik.
By annotating the digital corpus within the UD framework, we hope to contribute to expanding the framework to annotate other polysynthetic languages.
It would further allow us to utilize existing UD tools (e.g. multilingual UD parser) for comparative linguistic research as well as other NLP tasks like syntactic parsing.

A second goal for the UD treebank project is to better understand the syntactic properties of Yupik and to utilize such knowledge for future NLP tasks.
%
In particular, we can use the treebank to create novel sentences in Yupik, thereby augmenting existing textual data.
This would greatly assist those NLP tasks that require considerable quantities of data, such as neural language modeling.

In summary, the digital corpus can help us achieve a better understanding of Yupik morphology and syntax, which in turn, would result in the building of more robust computational models.
These computational models would then support the development of educational applications for Yupik revitalization, such as spell-checkers, text-completion systems, interactive e-books, and language learning apps.

\section{Future Work\label{sec:additional}}

To date, we have digitized all of the Yupik-language materials at the Gambell school and a portion of those archived at the Alaska Native Language Archive (ANLA).
There are a number of other materials located in the ANLA, however, that have not yet been included in the digital corpus.

After visiting the ANLA in early 2019, we have identified approximately 65 documents indexed under St.~Lawrence Island Yupik or Chaplinski Yupik that remain to be scanned.
We have also confirmed that there is a substantial amount of Yupik material at the ANLA that has neither been indexed nor scanned, most of which are Soviet-era Yupik texts (primarily in Cyrillic orthography) collected by Michael Krauss during visits to various libraries in the Soviet Union \cite{Krauss:1971}.

Furthermore, the Yupik examples in \newcite{Shinen:1982}, \newcite{Silook:1983}, \newcite{deReuse:1994}, \newcite{Shutt:2014} have not been digitized, nor have the examples in Soviet-era Yupik language documentation \cite{Menovshchikov:1960,Menovshchikov:1962,Menovshchikov:1967,Menovshchikov:1983}.
The latter are written in Cyrillic orthography with descriptions in Russian.
Future work will entail digitizing all of these materials.

A second objective for the digital corpus is text verification.
While ABBYY is the state-of-the-art software for OCR work, errors may still have occurred during the OCR process.
As such, we plan to have all digitized texts verified by native speakers.

Lastly, we intend to use the digital corpus to eventually build a \textit{parallel} corpus of Yupik texts and their translations.
Many of the texts included in the digital corpus have English translations, while many of the Soviet-era works that we plan to include have Russian translations.
One challenge, however, is the fact that many of the translations do not have a one-to-one correspondence with Yupik sentences.
In such cases, a single Yupik sentence may be translated as more than one English sentence, or vice versa.
The intended parallel corpus will map Yupik sentences one-to-one to their translations, which would facilitate various projects and endeavors in NLP.

\section{Conclusion}

The Yupik language and the corpus of Yupik written texts described herein represent important components of the linguistic and cultural heritage of the St.~Lawrence Island Yupik people.
While many of the existing Yupik-language texts have already been fully digitized and are present in our digital corpus, there remains much ongoing and future work.
As it stands, however, the digital corpus already lends itself to general linguistic inquiry and research related to NLP.
We believe it to be a valuable source of data that would greatly contribute to our understanding of the Yupik language, and moreover, to the field of NLP, as computational research on polysynthetic languages is still relatively scarce.
Above all, however, is the fact that the digital corpus has broadened the accessibility of Yupik language materials, which is a pivotal step towards establishing a program for Yupik language education and revitalization.

\section{Acknowledgments}

The Yupik language is a critical part of the cultural heritage of the Yupik people.
We offer our deep gratitude to the people of St.~Lawrence Island who have trusted us to work with this material.
Special thanks to the Yupik speakers whose words are recorded in this corpus.
We wish to thank everyone who assisted in scanning, proofreading, and digitizing this material.
Thanks to the board members and staff of the Native Village of Gambell, the City of Gambell, and Sivuqaq, Inc. Thanks to the Bering Strait School District, the faculty and staff of ANLC and ANLA, Dave and Mitzi Shinen of Wycliffe Bible Translators, Steven A. Jacobson, Kayo Nagai, Willem de Reuse, the staff of Gambell Lodge, Iyaaka (Anders Apassingok), Taayqa (Michael James, RIP) Rob Taylor, Petuwaq (Chris Koonooka) and the current and former faculty and staff of Gambell and Savoonga Schools who developed so many wonderful materials over the years and who supported us in this project.
This work was supported by NSF Awards \href{https://www.nsf.gov/awardsearch/showAward?AWD_ID=1761680}{1761680} and \href{https://www.nsf.gov/awardsearch/showAward?AWD_ID=1760977}{1760977}.

Igamsiqanaghhalek!

\bibliography{computel3}

\begin{thebibliography}{32}
\expandafter\ifx\csname natexlab\endcsname\relax\def\natexlab#1{#1}\fi

\bibitem[{Apassingok et~al.(1993)Apassingok, Uglowook, Koonooka, and
  Tennant}]{Apassingok:1993}
Anders Apassingok, (Iyaaka), Jessie Uglowook, (Ayuqliq), Lorena Koonooka,
  (Inyiyngaawen), and Edward Tennant, (Tengutkalek), editors. 1993.
\newblock \emph{Kallagneghet / {Drumbeats}}.
\newblock Bering Strait School District, Unalakleet, Alaska.

\bibitem[{Apassingok et~al.(1994)Apassingok, Uglowook, Koonooka, and
  Tennant}]{Apassingok:1994}
Anders Apassingok, (Iyaaka), Jessie Uglowook, (Ayuqliq), Lorena Koonooka,
  (Inyiyngaawen), and Edward Tennant, (Tengutkalek), editors. 1994.
\newblock \emph{Akiingqwaghneghet / {Echoes}}.
\newblock Bering Strait School District, Unalakleet, Alaska.

\bibitem[{Apassingok et~al.(1995)Apassingok, Uglowook, Koonooka, and
  Tennant}]{Apassingok:1995}
Anders Apassingok, (Iyaaka), Jessie Uglowook, (Ayuqliq), Lorena Koonooka,
  (Inyiyngaawen), and Edward Tennant, (Tengutkalek), editors. 1995.
\newblock \emph{Suluwet / {Whisperings}}.
\newblock Bering Strait School District, Unalakleet, Alaska.

\bibitem[{Apassingok et~al.(1985)Apassingok, Walunga, and
  Tennant}]{Apassingok:1985:Vol1}
Anders Apassingok, (Iyaaka), Willis Walunga, (Kepelgu), and Edward Tennant,
  (Tengutkalek), editors. 1985.
\newblock \emph{Sivuqam Nangaghnegha --- Siivanllemta Ungipaqellghat / {Lore}
  of {St. Lawrence Island} --- {Echoes} of our {Eskimo} Elders}, volume 1:
  Gambell.
\newblock Bering Strait School District, Unalakleet, Alaska.

\bibitem[{Apassingok et~al.(1987)Apassingok, Walunga, and
  Tennant}]{Apassingok:1987:Vol2}
Anders Apassingok, (Iyaaka), Willis Walunga, (Kepelgu), and Edward Tennant,
  (Tengutkalek), editors. 1987.
\newblock \emph{Sivuqam Nangaghnegha --- Siivanllemta Ungipaqellghat / {Lore}
  of {St. Lawrence Island} --- {Echoes} of our {Eskimo} Elders}, volume 2:
  Savoonga.
\newblock Bering Strait School District, Unalakleet, Alaska.

\bibitem[{Apassingok et~al.(1989)Apassingok, Walunga, and
  Tennant}]{Apassingok:1989:Vol3}
Anders Apassingok, (Iyaaka), Willis Walunga, (Kepelgu), and Edward Tennant,
  (Tengutkalek), editors. 1989.
\newblock \emph{Sivuqam Nangaghnegha --- Siivanllemta Ungipaqellghat / {Lore}
  of {St. Lawrence Island} --- {Echoes} of our {Eskimo} Elders}, volume 3:
  Southwest Cape.
\newblock Bering Strait School District, Unalakleet, Alaska.

\bibitem[{Badten et~al.(2008)Badten, Kaneshiro, Oovi, and
  Koonooka}]{Badten:2008}
Linda~Womkon Badten, Vera~Oovi Kaneshiro, Marie Oovi, and Christopher Koonooka.
  2008.
\newblock \emph{{St.~Lawrence Island / Siberian Yupik Eskimo} Dictionary}.
\newblock Alaska Native Language Center, University of Alaska Fairbanks.

\bibitem[{Chen et~al.(2020)Chen, Park, and Schwartz}]{Chen:2020}
Emily Chen, Hyunji~Hayley Park, and Lane Schwartz. 2020.
\newblock Improving finite-state morphological analysis for {St.~Lawrence
  Island Yupik} with paradigm function morphology.

\bibitem[{Chen and Schwartz(2018)}]{Chen:2018}
Emily Chen and Lane Schwartz. 2018.
\newblock A morphological analyzer for {St.~Lawrence Island} / {Central
  Siberian Yupik}.
\newblock In \emph{Proceedings of the 11th Language Resources and Evaluation
  Conference}, Miyazaki, Japan.

\bibitem[{Fortescue et~al.(2010)Fortescue, Jacobson, and
  Kaplan}]{fortescue-etal-2010}
Michael Fortescue, Steven Jacobson, and Lawrence Kaplan. 2010.
\newblock \emph{Comparative {E}skimo Dictionary with {A}leut Cognates}, 2nd
  edition.
\newblock Alaska Native Language Center, Fairbanks, Alaska.

\bibitem[{Jacobson(2001)}]{Jacobson:2001}
Steven~A. Jacobson. 2001.
\newblock \emph{A Practical Grammar of the {St.~Lawrence Island / Siberian
  Yupik Eskimo} Language, Preliminary Edition}, 2nd edition.
\newblock Alaska Native Language Center, Fairbanks, Alaska.

\bibitem[{Koonooka(2005)}]{Koonooka:2005}
Christopher Koonooka, (Petuwaq). 2005.
\newblock Yupik language instruction in {Gambell} ({St. Lawrence Island},
  {Alaska}).
\newblock \emph{Études/Inuit/Studies}, 29(1/2):251--266.

\bibitem[{Koonooka(2003)}]{Koonooka:2003}
Christopher~(Petuwaq) Koonooka. 2003.
\newblock \emph{Ungipaghaghlanga: Let Me Tell You A Story}.
\newblock Alaska Native Language Center.

\bibitem[{Krauss(1971)}]{Krauss:1971}
Michael Krauss. 1971.
\newblock Developing a literature in the language of the {Eskimos} of {St.
  Lawrence Island}.
\newblock Alaska Native Language Archive Identifier SY970K1971c.

\bibitem[{Krauss(1980)}]{Krauss:1980}
Michael Krauss. 1980.
\newblock Alaska {Native} languages: Past, present and future.
\newblock \emph{ANLC Research Papers}, 4.

\bibitem[{Krauss et~al.(2011)Krauss, Holton, Kerr, and West}]{Krauss:2011}
Michael Krauss, Gary Holton, Jim Kerr, and Colin~T. West. 2011.
\newblock Indigenous peoples and languages of {Alaska}.
\newblock Alaska Native Language Archive Identifier G961K2010.

\bibitem[{Krupnik and Chlenov(2013)}]{Krupnik:2013}
Igor Krupnik and Michael Chlenov. 2013.
\newblock \emph{Yupik Transitions --- {Change} and Survival at {Bering}
  {Strait}, 1900-1960}.
\newblock University of Alaska Press, Fairbanks, Alaska.

\bibitem[{Menovshchikov(1960)}]{Menovshchikov:1960}
G.~A. Menovshchikov. 1960.
\newblock \emph{Eskimosskii iazyk}.
\newblock Gosudarstvennoe uchebno-pedagogicheskoe izdatel'stvo, Leningrad.
\newblock Pedagogical grammar, similar in scope and level to Jacobson (2001).

\bibitem[{Menovshchikov(1962)}]{Menovshchikov:1962}
G.~A. Menovshchikov. 1962.
\newblock \emph{Grammatika iazyka aziatskikh eskimosov (Grammar of the language
  of Asian Eskimos)}, volume~1.
\newblock Izdatel'stvo akademii Nauk (Academy of Sciences of the USSR), Moscow
  and Leningrad.

\bibitem[{Menovshchikov(1967)}]{Menovshchikov:1967}
G.A. Menovshchikov. 1967.
\newblock \emph{Grammatika iazyka aziatskikh eskimosov}, volume~2.
\newblock Izdatel'stvo akademii Nauk, Moscow and Leningrad.
\newblock Major reference grammar.

\bibitem[{Menovshchikov(1983)}]{Menovshchikov:1983}
G.A. Menovshchikov. 1983.
\newblock \emph{Slovar' ekimossko-russkiy i russkio-eskimosskiy.}
\newblock Prosveshchenie., Leningrad.
\newblock Yupik to Russian and Russian to Yupik School dictionary.

\bibitem[{Morgounova(2007)}]{Morgounova:2007}
Daria Morgounova. 2007.
\newblock Language, identities and ideologies of the past and present
  {Chukotka}.
\newblock \emph{Études/Inuit/Studies}, 31(1-2):183--200.

\bibitem[{Nagai(2001)}]{Nagai:2001}
Kayo Nagai. 2001.
\newblock \emph{{Mrs. Della Waghiyi's} {St. Lawrence Island Yupik} Texts with
  Grammatical Analysis}.
\newblock Number A2-006 in Endangered Languages of the Pacific Rim. Nakanishi
  Printing, Kyoto, Japan.

\bibitem[{Nivre et~al.(2016)Nivre, de~Marneffe, Ginter, Goldberg, Haji{\v{c}},
  Manning, McDonald, Petrov, Pyysalo, Silveira, Tsarfaty, and
  Zeman}]{nivre-etal-2016-universal}
Joakim Nivre, Marie-Catherine de~Marneffe, Filip Ginter, Yoav Goldberg, Jan
  Haji{\v{c}}, Christopher~D. Manning, Ryan McDonald, Slav Petrov, Sampo
  Pyysalo, Natalia Silveira, Reut Tsarfaty, and Daniel Zeman. 2016.
\newblock {U}niversal {D}ependencies v1: A multilingual treebank collection.
\newblock In \emph{Proceedings of the Tenth International Conference on
  Language Resources and Evaluation ({LREC}'16)}, pages 1659--1666,
  Portoro{\v{z}}, Slovenia. European Language Resources Association (ELRA).

\bibitem[{{Nome Agency Bilingual Education Resource
  Center}(1975)}]{Aghnaaneq_Neghsaq_Teghigniqelghii}
{Nome Agency Bilingual Education Resource Center}. 1975.
\newblock \emph{Aghnaaneq Neghsaq Teghigniqelghii}.
\newblock GSA Printing Plant, Juneau, AK.

\bibitem[{de~Reuse(1994)}]{deReuse:1994}
Willem~J. de~Reuse. 1994.
\newblock \emph{{Siberian} {Yupik} {Eskimo} --- The Language and Its Contacts
  with {Chukchi}}.
\newblock Studies in Indigenous Languages of the Americas. University of Utah
  Press, Salt Lake City, Utah.

\bibitem[{Shinen(1982)}]{Shinen:1982}
David~C. Shinen. 1982.
\newblock Some beginning conversational phrases in {St. Lawrence Island Yupik
  Eskimo}.
\newblock Alaska Native Language Center Identifier SY960S1982.

\bibitem[{Shutt et~al.(2014)Shutt, Biddison, and Koonooka}]{Shutt:2014}
Lauren Shutt, Dawn Biddison, and Christopher Koonooka. 2014.
\newblock \emph{Listen \& Learn: St. Lawrence Island Yupik Language and Culture
  Video Lessons}.
\newblock Arctic Studies Center, Smithsonian Institution, Anchorage, Alaska.

\bibitem[{Silook et~al.(1983)Silook, Badten, Carius, Kaneshiro, Oozeva, Shinen,
  and Slwooko}]{Silook:1983}
Roger Silook, Adelinda~Womkon Badten, Helen~Slwooko Carius, Vera~Oovi
  Kaneshiro, Elinor Oozeva, David Shinen, and Grace Slwooko. 1983.
\newblock \emph{Sivuqam Anglinghhaan Akuzisii / St. Lawrence Island Junior
  Dictionary.}
\newblock National Bilingual Materials Development Center, Rural Education
  Affairs, University of Alaska., Anchorage, Alaska.
\newblock Yupik-to-English school dictionary.

\bibitem[{Slwooko(1977)}]{Slwooko:1977}
Grace Slwooko. 1977.
\newblock \emph{Sivuqam Ungipaghaatangi I}.
\newblock University of Alaska, Anchorage, AK.

\bibitem[{Slwooko(1979)}]{Slwooko:1979}
Grace Slwooko. 1979.
\newblock \emph{Sivuqam Ungipaghaatangi II}.
\newblock University of Alaska, Anchorage, AK.

\bibitem[{Wycliffe(2018)}]{YupikBible}
Wycliffe. 2018.
\newblock \emph{Yupik {N}ew {T}estament}.
\newblock Wycliffe Bible Translators, Saint Lawrence Island, AK.

\end{thebibliography}
\bibliographystyle{acl_natbib_nourl}

\end{document}